# Applications of fuzzy logic to Case-Based Reasoning

Igor Ya. Subbotin, Michael Gr. Voskoglou

Professor of Mathematics,
College of Letters and Sciences,
National University, Los Angeles, California, USA
e-mail: isubboti@nu.edu

Professor of Mathematical Sciences
School of Technological Applications
Graduate Technological Educational Institute, Patras, Greece
e-mail:  mvosk@hol.gr , voskoglou@teipat.gr

*In memoriam Filippo Spagnolo*

### Abstract

*The article discusses some applications of fuzzy logic ideas to formalizing of the Case-Based Reasoning (CBR) process and to measuring the effectiveness of CBR systems*

## 1     Introduction

Broadly construed *Case-Based Reasoning (CBR)* is the process of solving new problems based on the solution of past problems. The CBR systems' expertise is embodied in a collection (library) of past cases rather, than being encoded in classical rules. Each case typically contains a description of the problem plus a solution and/or the outcomes. When a problem is successfully solved, the experience is retained in order to solve similar problems in future. When an attempt to solve a problem fails, the reason for the failure is identified and remembered in order to avoid the same mistake in future. Thus CBR is a cyclic and integrated process of solving a problem, learning from this experience, solving a new problem, etc. A case-library can be a powerful corporate resource allowing everyone in an organization to tap in the corporate library, when handling a new problem. CBR allows the case-library to be developed incrementally, while its maintenance is relatively easy and can be carried out by domain experts. As an intelligent-systems' method CBR enables information managers to increase efficiency and reduce cost by substantially automating processes such as diagnosis, scheduling and design.

CBR has been formalized for purposes of computer and human reasoning as a four steps process. These steps involve:

**$R_1$:** *Retrieve* the most similar to the new problem past case.

**$R_2$:** *Reuse* the information and knowledge of the retrieved case for the solution of the new problem.

**$R_3$:** *Revise* the proposed solution.

**$R_4$:** *Retain* the part of this experience likely to be useful for future problem-solving.

Riesbeck and Bain [11], Slade [12], Lei et al. [9], Aamodt and Plaza [1], Voskoglou ([17], [20]), etc have provided detailed flowcharts illustrating the steps of the CBR process.

## 2. Voskoglou's Fuzzy model for CBR

Created by Zadeh [24], fuzzy logic has been successfully developed by many researchers and has been proven to be extremely productive in many applications (see, for example, [2], [6], [7], [18], [21], [22], [23] and others). There are also some interesting attempts to implement fuzzy logic ideas in the field of education ([4], [9], [10], [13], [16], [19], [22], etc).

Voskoglou in the articles [18] and [21] has developed a fuzzy set model for describing a CBR system. In the following few paragraphs we cite parts of these articles.

"Let us consider a CBR system whose library contains n past cases, $n \geq 2$. We denote by $R_i$, i=1,2,3, the steps of retrieval, reuse and revision and by a, b, c, d, and e the linguistic labels of negligible, low, intermediate, high and complete degree of success respectively for each of the $R_i$'s. Set

$$U=\{a, b, c, d, e\}$$

We are going to represent $R_i$'s as fuzzy sets in U. For this, if $n_{ie}$, $n_{id}$, $n_{ic}$, $n_{ib}$ and $n_{ia}$ respectively denote the number of cases where it has been achieved negligible, low, intermediate, high and complete degree of success for the state $R_i$ i=1,2,3, we define the membership function $m_{Ri}$ in terms of the frequencies, i.e. by

$$m_{Ri}(x) = \frac{n_{ix}}{n}$$

for each x in U. Thus we can write

$$R_i = \{(x, \frac{n_{ix}}{n}) : x \in U\}, i=1,2,3$$

The reason, for which we didn't include the last step $R_4$ of the CBR process in our fuzzy representation, is that all past cases, either successful, or not, are

Some Applications of Fuzzy Logic…

retained in the system's library and therefore there is no fuzziness in this case. In other words keeping the same notation we have that $n_{4a}=n_{4b}=n_{4c}=n_{4d}=0$ and $n_{4e}=1$.
In order to represent all possible *profiles (overall states)* of a case during the CBR process, we consider a fuzzy relation, say R, in $U^3$ of the form

$$R=\{(s, m_R(s)) : s=(x, y, z) \in U^3\}$$

To determine properly the membership function $m_R$ we give the following definition:

*A triple is said to be well ordered if x corresponds to a degree of success equal or greater than y, and y corresponds to a degree of acquisition equal or greater than z.*

For example, the profile (c, c, a) is well ordered, while (b, a, c) is not.
We define now the membership degree of s to be

$$m_R(s)=m_{R_1}(x)m_{R_2}(y)m_{R_3}(z)$$

if s is a well ordered profile, and zero otherwise. In fact, if for example (b, a, c) possessed a nonzero membership degree, given that the degree of success at the step of reuse is negligible how the proposed solution could be revised?
In order to simplify our notation we shall write $m_s$ instead of $m_R(s)$. Then the *possibility* $r_s$ of the profile s is given by

$$r_s = \frac{m_s}{\max\{m_s\}}$$

where $\max\{m_s\}$ denotes the maximal value of $m_s$, for all s in $U^3$. In other words $r_s$ is the "relative membership degree" of s with respect to the other profiles".
Further, Voskoglou ( [18], [21]) argues that the *total possibilistic uncertainty* T(r) (i.e. the sum of *strife* and *non specificity* [8, p.28]) on the ordered possibility distribution r of the profiles of a CBR system can be used as a measure of its efficiency in solving problems related to its cases. In fact, the amount of information obtained by an action can be measured by the reduction of uncertainty resulting from this action. Accordingly system's uncertainty during the CBR process is connected to its capacity in obtaining relevant information. The lower is T(r) (which means grater reduction of the system's uncertainty) the greater the system's efficiency in solving related problems.
In order to illustrate the use of his model in practice Voskoglou [18] presented the following EXAMPLE:
"Let us consider a CBR system with an existing library of 105 past cases, where in no case there was a failure at the step of retrieval of a past case for the solution of the corresponding problem. More explicitly, let us assume that in 51 cases we had an intermediate success in retrieving a suitable past case, in 24 cases high, and in 30 cases we had a complete success respectively. Of course the existence of a certain criterion is necessary in order to be able to characterize the degree of success of retrieval for each of the past cases. Thus the step of retrieval can be represented as a fuzzy set in U as

$$R_1 = \{(a,0),(b,0),(c, \tfrac{51}{105}),(d, \tfrac{24}{105}),(e, \tfrac{30}{105})\}.$$

Assume further that in a similar way we obtained that
$$R_2 = \{(a, \tfrac{18}{105}),(b, \tfrac{18}{105}),(c, \tfrac{48}{105}),(d, \tfrac{21}{105}),(e,0)\},$$
and
$$R_3 = \{(a, \tfrac{36}{105}),(b, \tfrac{30}{105}),(c, \tfrac{39}{105}),(d,0),(e,0)\}.$$

It is a straightforward process now to calculate the membership degrees of all the possible profiles (see [18]; column of $m_s(1)$ in Table 1). For example, if $s=(c, b, a)$, then

$$m_s = m_{R_1}(c) \cdot m_{R_2}(b) \cdot m_{R_3}(a) = \tfrac{51}{105} \cdot \tfrac{18}{105} \cdot \tfrac{36}{105} \approx 0{,}029.$$

It turns out that $(c, c, c)$ is the profile with the maximal membership degree 0,082 and therefore the possibility of each $s$ in $U^3$ is given by $r_s = \tfrac{m_s}{0{,}082}$. Calculating the possibilities of the $5^3=125$ (ordered samples with replacement of 3 objects out of 5) in total profiles Voskoglou found that the total possibilistic uncertainty of the system is 2,97.

Next he considered another CBR system, designed for the solution of the same type of problems, with an existing library of 90 past cases and working as before he found that
$$R_1 = \{(a,0),(b, \tfrac{18}{90}),(c, \tfrac{45}{90}),(d, \tfrac{27}{90}),(e,0)\},$$

$$R_2 = \{(a, \tfrac{18}{90}),(b, \tfrac{24}{90}),(c, \tfrac{48}{90}),(d, 0),(e,0)\},$$
and
$$R_3 = \{(a, \tfrac{36}{90}),(b, \tfrac{27}{90}),(c, \tfrac{27}{90}),(d,0),(e,0)\}.$$

From the calculation of all possible profiles it turns out that $(c, c, a)$ is the profile possessing the maximal membership degree 0,107 and therefore the possibility of each $s$ is given by $r_s = \tfrac{m_s}{0{,}107}$

Calculating the possibilities of all profiles Voskoglou ([18], [21]) found that $T(r)=2{,}322$

Thus, since $2{,}322 < 2{,}97$ the effectiveness of the second system in solving new related problems is better than that of the first one.

Notice that in general, the more are the stored past cases in the system's library, the greater is expected to be its effectiveness in solving new related problems. In fact, the more are the past cases, the greater is the probability for a new problem to fit satisfactorily to one of them. Therefore the fact that the second system was found to be more effective than the first one, although not impossible to happen, it is rather unexpected in general.

## 3   Application of the Subbotin's model

The following model employs a different approach to a comprehensive assessment. The main base of this approach has been developed in [13]. This approach is visible, does not implement any complicated calculations on the final step, and, what is important, can be employed to a single case's assessment and to the system's assessment as well.

In the fuzzy systems, there is a commonly used approach to measure the performance by graphically representing the information as a two dimensional figure $F$ and work with coordinates of the center of mass $F_c(x_c, y_c)$ of this figure (see for example, ([3], [5], [15]).

We can calculate it using the following well-known formulas:

$$(1) \quad x_c = \frac{\iint_F x\,dxdy}{\iint_F dxdy}, \quad y_c = \frac{\iint_F y\,dxdy}{\iint_F dxdy}.$$

As any assessment, our approach is very approximate. So it would be much more useful in everyday life to simplify the situation assuming that our figure approximated with bar graph like on the following Figure 1.

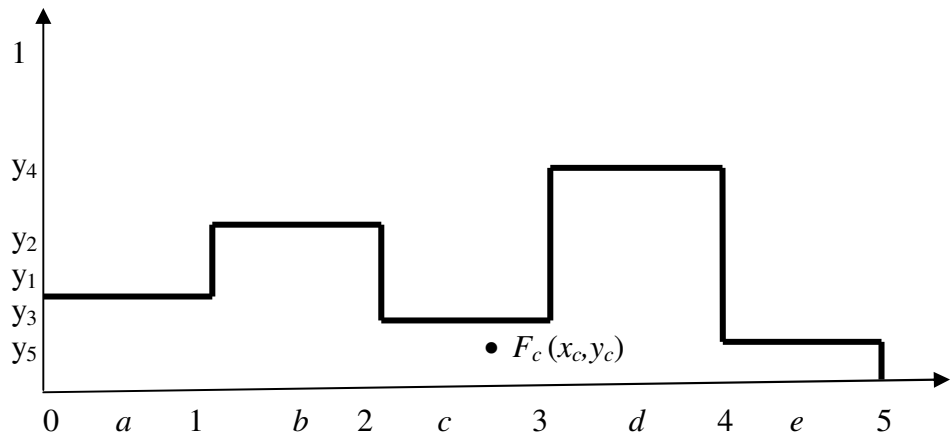

Figure 1:  Bar graphical data representation

It is easy to see that in the case when our figure consists of $n$ rectangles, the formulas (1) can be reduced to the following formulas [13]:

$$(2) \quad x_c = \frac{1}{2}\left(\frac{\sum_{i=1}^{n}(2i-1)y_i}{\sum_{i=1}^{n} y_i}\right), \quad y_c = \frac{1}{2}\left(\frac{\sum_{i=1}^{n} y_i^2}{\sum_{i=1}^{n} y_i}\right).$$

Let us consider the mentioned above CBR system whose library contains n past cases, n≥2. We denote by $R_i$, i=1,2,3, the steps of retrieval, reuse and revision and by *a, b, c, d,* and *e* the linguistic labels of negligible, low, intermediate, high and complete degree of success respectively for each of the $R_i$'s. We can measure the effectiveness using the following numerical point distribution: $\alpha \in [0,1)$, $b \in [1,2)$, $c \in [2,3)$, $d \in [3,4)$ and $e \in [4,5]$.

Now formulas (2) will be transformed into the following formulas:

$$x_c = \frac{1}{2}\left(\frac{y_1 + 3y_2 + 5y_3 + 7y_4 + 9y_5}{y_1 + y_2 + y_3 + y_4 + y_5}\right),$$

$$y_c = \frac{1}{2}\left(\frac{y_1^2 + y_2^2 + y_3^2 + y_4^2 + y_5^2}{y_1 + y_2 + y_3 + y_4 + y_5}\right).$$

Since we can assume that

$$y_1 + y_2 + y_3 + y_4 + y_5 = 1,$$

we can write

$$(3) \quad \begin{aligned} x_c &= \frac{1}{2}(y_1 + 3y_2 + 5y_3 + 7y_4 + 9y_5), \\ y_c &= \frac{1}{2}(y_1^2 + y_2^2 + y_3^2 + y_4^2 + y_5^2) \end{aligned}$$

where $y_i$, $1 \leq i \leq 5$, is the ratio of the cases in the system having the labels *a, b, c, d,* and *e* to the numbers of all cases in the system.

With the help of some elementary inequalities it is not difficult to establish that the unique minimum is reached at the point $F_m(2.5, \frac{1}{10})$ when

$y_1 = y_2 = y_3 = y_4 = y_5 = \frac{1}{5}$.

Indeed, since

$$y_1 + y_2 + y_3 + y_4 + y_5 = 1,$$
$$(y_1 + y_2 + y_3 + y_4 + y_5)^2 = 1.$$

Therefore

$$(y_1 + y_2 + y_3 + y_4 + y_5)^2 = y_1^2 + y_2^2 + y_3^2 + y_4^2 + y_5^2 +$$
$$+ 2y_1 y_2 + 2y_1 y_3 + 2y_1 y_4 + 2y_1 y_5 + 2y_2 y_3 + 2y_2 y_4 + 2y_2 y_5 +$$
$$+ 2y_3 y_4 + 2y_3 y_5 + 2y_4 y_5 \leq y_1^2 + y_2^2 + y_3^2 + y_4^2 + y_5^2 +$$
$$y_1^2 + y_2^2 + y_1^2 + y_3^2 + y_1^2 + y_4^2 + y_1^2 + y_5^2 + y_2^2 + y_3^2 +$$
$$y_2^2 + y_4^2 + y_2^2 + y_5^2 + y_3^2 + y_4^2 + y_3^2 + y_5^2 + y_4^2 + y_5^2 =$$
$$= 5(y_1^2 + y_2^2 + y_3^2 + y_4^2 + y_5^2)$$

where the equality holds in the case when

$$y_1 = y_2 = y_3 = y_2 = y_5 = \frac{1}{5}.$$

In this case,

$$x_c = \frac{1}{2}(y_1 + 3y_2 + 5y_3 + 7y_4 + 9y_5) = 2.5.$$

The ideal case is when $y_1=y_2=y_3=y_4=0$ and $y_5=1$. Then from formulas (3) we get that $x_c = \frac{9}{2}$ and $y_c = \frac{1}{2}$. Therefore the center of mass in this case is the point $F_i$ $(\frac{9}{2}, \frac{1}{2})$.

On the other hand the worst case is when $y_1=1$ and $y_2=y_3=y_4=y_5=0$. Then for formulas (3) we find that the center of mass is the point $F_w$ $(\frac{1}{2}, \frac{1}{2})$.

Now we can formulate our criterion for comparing the two groups' performances in the following form (for more details see [14]) :

*Among two or more groups the group with the biggest $x_c$ performs better;*
(4) *If two or more groups have the same $x_c \geq 2.5$, then the group with the higher $y_c$ performs better. If two or more groups have the same $x_c \leq 2.5$, then the group with the lower $y_c$ performs better.*

In the experiment illustrated the Voskoglou's model above, the step of retrieval can be represented as the following

$$R_{11} = \{(a,0),(b,0),(c, \tfrac{51}{105}),(d, \tfrac{24}{105}),(e, \tfrac{30}{105})\}.$$

Assume further that in a similar way we obtained that

$$R_{12} = \{(a, \tfrac{18}{105}), (b, \tfrac{18}{105}), (c, \tfrac{48}{105}), (d, \tfrac{21}{105}), (e, 0)\},$$

and

$$R_{13} = \{(a, \tfrac{36}{105}), (b, \tfrac{30}{105}), (c, \tfrac{39}{105}), (d, 0), (e, 0)\}.$$

Next we considered another CBR system, designed for the solution of the same type of problems, with an existing library we found that

$$R_{21} = \{(a, 0), (b, \tfrac{18}{90}), (c, \tfrac{45}{90}), (d, \tfrac{27}{90}), (e, 0)\},$$

$$R_{22} = \{(a, \tfrac{18}{90}), (b, \tfrac{24}{90}), (c, \tfrac{48}{90}), (d, 0), (e, 0)\},$$

and

$$R_{23} = \{(a, \tfrac{36}{90}), (b, \tfrac{27}{90}), (c, \tfrac{27}{90}), (d, 0), (e, 0)\}.$$

In the step one we have

$$x_{c11} = \frac{1}{2}\left(5 \cdot \frac{51}{105} + 7 \cdot \frac{24}{105} + 9 \cdot \frac{30}{105}\right) =$$

$$= \frac{1}{2}\left(\frac{255}{105} + \frac{168}{105} + \frac{270}{105}\right) = \frac{1}{2} \cdot \frac{693}{105} = 3.3,$$

$$y_{c11} = \frac{1}{2}\left(\frac{51^2}{105} + \frac{24^2}{105} + \frac{30^2}{105}\right) =$$

$$= \frac{1}{2}\left(\frac{2601}{11025} + \frac{576}{11025} + \frac{900}{11025}\right) = \frac{1}{2}\left(\frac{4077}{11025}\right) \approx 0.185.$$

$$x_{c21} = \frac{1}{2}\left(3 \cdot \frac{18}{90} + 5 \cdot \frac{45}{90} + 7 \cdot \frac{27}{90}\right) =$$

$$= \frac{1}{2}\left(\frac{54}{90} + \frac{225}{90} + \frac{189}{90}\right) = \frac{1}{2}\left(\frac{468}{90}\right) = 2.6,$$

$$y_{c21} = \frac{1}{2}\left(\frac{18^2}{90} + \frac{45^2}{90} + \frac{27^2}{90}\right) =$$

$$\frac{1}{2}\left(\frac{324}{8100} + \frac{2025}{8100} + \frac{729}{8100}\right) =$$

$$= \frac{1}{2}\left(\frac{3078}{8100}\right) = 0.19.$$

By the criterion (4), the first group demonstrates significantly better performance.

In the step two, we have

$$R_{12} = \{(a, \tfrac{18}{105}), (b, \tfrac{18}{105}), (c, \tfrac{48}{105}), (d, \tfrac{21}{105}), (e, 0)\},$$

$$R_{22} = \{(a, \tfrac{18}{90}), (b, \tfrac{24}{90}), (c, \tfrac{48}{90}), (d, 0), (e, 0)\},$$

and respectively

$$x_{c12} = \frac{1}{2}\left(\frac{18}{105} + 3 \cdot \frac{18}{105} + 5 \cdot \frac{48}{105} + 7 \cdot \frac{21}{105}\right) =$$

$$= \frac{1}{2}\left(\frac{459}{105}\right) \approx 2.186,$$

$$y_{c12} = \frac{1}{2}\left(\frac{18^2}{105} + \frac{18^2}{105} + \frac{48^2}{105} + \frac{21^2}{105}\right) =$$

$$= \frac{1}{2}\left(\frac{3393}{11025}\right) \approx 0.154.$$

$$x_{c22} = \frac{1}{2}\left(\frac{18}{90} + 3 \cdot \frac{24}{90} + 5 \cdot \frac{48}{90}\right) = \frac{1}{2}\left(\frac{18}{90} + \frac{72}{90} + \frac{240}{90}\right)$$

$$= \frac{1}{2}\left(\frac{330}{90}\right) \approx 1.833,$$

$$y_{c22} = \frac{1}{2}\left(\frac{18^2}{90} + \frac{24^2}{90} + \frac{48^2}{90}\right) = \frac{1}{2}\left(\frac{3204}{8100}\right) \approx 0.198.$$

By the criterion (4), the first group again demonstrates better performance.

And in the final third step we have

$$R_{13} = \{(a, \tfrac{36}{105}), (b, \tfrac{30}{105}), (c, \tfrac{39}{105}), (d, 0), (e, 0)\},$$

$$R_{23} = \{(a, \tfrac{36}{90}), (b, \tfrac{27}{90}), (c, \tfrac{27}{90}), (d, 0), (e, 0)\},$$

and respectively

$$x_{c13} = \frac{1}{2}\left(\frac{36}{105} + 3 \cdot \frac{30}{105} + 5 \cdot \frac{39}{105}\right) = \frac{1}{2}\left(\frac{321}{105}\right) \approx 1.529,$$

$$y_{c13} = \frac{1}{2}\left(\frac{36^2}{105} + \frac{30^2}{105} + \frac{39^2}{105}\right) = \frac{1}{2}\left(\frac{3717}{11025}\right) \approx 0.169$$

$$x_{c23} = \frac{1}{2}\left(\frac{36}{90} + 3 \cdot \frac{27}{90} + 5 \cdot \frac{27}{90}\right) = \frac{1}{2}\left(\frac{252}{90}\right) = 1.4,$$

$$y_{c13} = \frac{1}{2}\left(\frac{36^2}{90} + \frac{27^2}{90} + \frac{27^2}{90}\right) = \frac{1}{2}\left(\frac{2754}{8100}\right) = 0.17$$

So in this step, the performances of both groups are close, but the first group performs slightly better..

Based on our calculation we can conclude that the first group demonstrated better at all three steps. We can also compare each group performance at each step. Both groups performed better at the first step, and the worse at the third step. This directly reflects the ascending complication of the tasks at the second step and especially the third step.

## 4. Conclusions

CBR is one of the central ideas in the nowadays artificial intelligence. Its applications are especially efficient in helping information managers to increase efficiency and reduce cost by substantially automating processes. Applying fuzzy logic to formalization of the CBR process helps in obtaining quantitative information for this process (comparing systems' performances, etc), as well as a qualitative view on the degree of success at the successive steps of the CBR process through the calculation of the possibilities of all system's profiles. The described in the article fuzzy models help the users to get specific concrete information regarding the existing CBR systems and to choose the appropriate one for the solution of their problems.

We emphasize the fact that these two models are treating differently the idea of a CBR system's performance. In fact, in Voskoglou's model the system's uncertainty during the CBR process is connected to its capacity in obtaining the relevant information. Under this sense, the lower is the system's total possibilitic uncertainty (which means greater reduction of the initially existing uncertainty), the better is its performance. On the other hand, in Subbotin's model the weighted average plays the main role, i.e. the result of the performance close to the ideal performance have much more weight than the one close to the lower end. In other words, while the first model is looking to the average performance, the second one is mostly looking at the quality of the performance. Therefore some differences could appear in boundary cases. This explains why, in the example presented in this paper, according to Voskoglou's model the first system was found to have a better performance than the second one, while just the opposite happened according to Subbotin's model.

In concluding, it is argued that a combined use of the two models helps in founding the ideal profile of performance according to the user's personal criteria of goals and therefore to choose the use of the appropriate CBR system, among the existing ones, for solving his/her problem.